\documentclass[pmlr]{jmlr}

\usepackage{longtable}

\usepackage{booktabs}
\usepackage[load-configurations=version-1]{siunitx} 

\theorembodyfont{\upshape}
\theoremheaderfont{\scshape}
\theorempostheader{:}
\theoremsep{\newline}

\jmlrvolume{299}
\jmlryear{2025}
\jmlrworkshop{Conference on Applied Machine Learning for Information Security}

\title[Evaluating Learning Harnesses Without Labels]{Evaluating Agentic Learning Harness Capabilities Without Labels via the Scaling Hypothesis}

\author{%
  \Name{Aryan Luthra} \Email{aryan@sublimesecurity.com}\\
  \Name{Kshitij Jain} \Email{kshitij@georgian.io}\\
  \Name{Siddharth Arya} \Email{siddharth.arya@georgian.io}\\
  \Name{Bobby Filar} \Email{bobby@sublimesecurity.com}\\
  \Name{Anna Bertiger} \Email{anna.b@sublimesecurity.com}
}

\usepackage{graphicx}
\usepackage{textcomp}
\usepackage{xcolor}
\usepackage{tikz}
\usepackage{listings}
\usepackage{subcaption}
\usepackage{xcolor}

\usepackage{pdflscape}
\usepackage{afterpage}
\usepackage{booktabs}
\usepackage{array}
\usepackage{longtable}
\usepackage{inconsolata}
\usepackage{booktabs}  
\usepackage{caption}   
\usepackage[authormarkup=none]{changes}
\usepackage{comment}

\definecolor{codegreen}{rgb}{0,0.6,0}
\definecolor{codegray}{rgb}{0.5,0.5,0.5}
\definecolor{codepurple}{rgb}{0.58,0,0.82}
\definecolor{backcolour}{rgb}{0.95,0.95,0.92}

\lstdefinelanguage{MQL}{
  keywords={type, disposition, sender, body, file, any, all, of, from, where, and, or, not, map, filter,
            length, any, all, flatten, to_string, lower, matches, starts_with, ends_with, contains},
  keywordstyle=\color{blue}\bfseries,
  comment=[l]{//}, 
  commentstyle=\color{codegreen},
  stringstyle=\color{codepurple},
  sensitive=false, 
  morecomment=[s]{/*}{*/}, 
  morestring=[b]", 
}

\begin{document}

\maketitle

\begin{abstract}
Agentic ``Continual Learning Harnesses", systems that pair an LLM with retrieval or memory to improve from feedback without retraining, have shown growing value in cybersecurity. But their value is conventionally measured by gains against labeled benchmarks, an approach that often fails in operational security settings. Benchmark labels are scarce, stale, and unrepresentatively sampled, so a practitioner often cannot tell whether a given harness helps at all, or which of two is better for their task. Traditional LLM-as-a-judge offers little signal because it is no stronger than the agent it evaluates, and distillation is unreliable on scarce, sporadic, and biased labels.

We propose a framework for evaluating learning harnesses end to end without a labeled benchmark, grounded in the scaling hypothesis. A stronger teacher model provides sparsely sampled corrections to a smaller student with a continual learning harness. We score a harness by how much its student converges toward the teacher over time. Across security tasks, model families, and harness designs, we show that improvement relative to the teacher correlates with improvement relative to a held-out gold standard, validating teacher-relative lift as a proxy for true harness uplift when labels are absent. We further show that LLM-as-a-judge between similarly powered models yields no usable signal. These results suggest that a teacher-sized model can be improved through the same harness when humans provide the same kind of sparse, high precision corrections.
\end{abstract}


\section{Introduction}

Agentic harnesses have grown to become the standard way to augment LLM based systems for real world tasks. Specifically in security tasks, where adversaries shift extremely fast, ``continual learning harnesses" are especially appealing. In this paper, we define a ``continual learning harness" as a system that wraps an LLM and takes in corrections over time in an attempt to improve the LLMs performance/personalization over time. An LLM with a harness that could rapidly adapt to incoming attacks is the perfect defender's tool. However, this same adaptability property makes these harnesses hard to evaluate, especially in operational security settings. Traditionally you would run a side by side benchmark and score the raw LLM, as well as the LLM plus 2-3 different harnesses and compare results on a labeled dataset. In many real security settings, there aren't enough labels to get a clear signal, and they are often biased to only the mistakes the models make on the hardest of questions. 

For example, Sublime Security has an AI agent named ASA (short for Autonomous Security Analyst), which analyzes email to determine if it is malicious, benign, graymail, or spam. This agent is equipped with tools and rubrics for deciding  an email's verdict and reporting on reasons for the decision. For most customers, the agent works autonomously, blocking thousands of emails a day. Feedback only comes in from customers when ASA is incorrect and has caused some kind of incident with significant consequences or visibility. This means that corrections and feedback are produced rarely, and sporadically, but with high precision. In addition, feedback is almost never produced when ASA is correct. Worse,  only the messages a customer explicitly shares reach Sublime, shrinking an already sparse signal further. So in order to build a harness to correct and personalize ASA from human feedback over time, we must derive a way to compare a harness' value without an explicitly labeled benchmark set. 

The traditional tools for optimization and measurement in continual learning situations both fail here. The first tool is LLM-as-a-judge: score the difference in performance for model with and without a harness by asking a second LLM which of their outputs are better. But operational security systems are often already running the strongest model available, so there is no stronger model left to serve as judge. Additionally, as we demonstrate, a same-strength judge is unable to recognize performance above its own baseline. The second tool is distillation: take the human corrections and bake them into the weights through fine-tuning or a lightweight adapter such as LoRA. But the corrections we have are too few, too sporadic, and too irregularly formed to fine-tune on. We are therefore left needing to build a learning harness and a way to tell whether it works, using almost no labeled data.

In this paper, we propose a new framework for measuring such learning harnesses without any labels. The framework relies on a teacher model and a student model, where the teacher significantly outperforms the student on the specific task being evaluated. Without labels, such a pair is hard to identify directly. We resolve this by appealing to the scaling hypothesis: we take two models from the same family, one larger and one smaller, and ask both to label a large pool of unlabeled samples for the task. We then look for substantial disagreement between them. Where the two disagree often, the scaling hypothesis gives us reason to believe the larger model is more often correct than the smaller one, since the two share a family and differ primarily in scale. This lets us designate the larger model as teacher and the smaller as student without ever observing a true label.

We then place the student in a continual learning harness and call the resulting model student*. Using the teacher's predictions as pseudo-labels, we supply sparse corrections to student* and measure the difference in performance between student and student* on a held-out set that the teacher has likewise pseudo-labeled. If the harness produces an uptick in performance under this purely pseudo-labeled regime, we take it as evidence that a model of capability similar to the teacher would see comparable gains from the same harness when given sparse corrections from real human labels.

We illustrate this utility in cases where we do have labeled data by comparing improvements of a learning harness using a smaller LLM student model and trained by a larger LLM teacher model with improvements relative to an oracle of correct labels. We show that across security tasks, model families and harnesses, improvements in performance relative to a teacher correlate with improvements in performance relative to a golden labeled data set. In contrast, a same strength LLM as judge is unable to recognize predictions better than its own baseline, and thus is not a viable strategy to assess improvements over a base model.

\section{Related Work}

\subsection{AI Security Evaluations}
Security evaluations, even without AI involved, can be difficult. Data is often proprietary and labels are often scarce, and become stale nearly as quickly as they are created. Add that to the difficulty in evaluating complex learning harnesses and agents, and the combination can be nearly impossible. 

There have been evaluations of agents for specific tasks where comparison to human results for the same task is possible, even if the data is not public, such as \cite{bertiger2025evaluating} which evaluates agent written detection rules in detection rates and rule quality, and \cite{bulut2026avda} measures quality by comparing to human written code using largely semantic comparisons. 

Within the cyber security research community, substantial efforts have been made to build increasingly capable benchmarks. Phishfuzzer \cite{toth2025phishfuzzer} creates realistic looking phish messages,AthenaBench \cite{alam2025athenabench} is an extension of CTIBench \cite{alam2024ctibench}, that leverages real-time cyber threat intelligence (CTI) data sources, such as MITRE ATT\&CK and the NVD API, to continuously generate nascent benchmark samples, ensuring models are evaluated against emerging vulnerabilities. AttackSeqBench \cite{ma2025attackseqbench} stages evaluation around multi-step reasoning tasks in the CTI space. SIABENCH \cite{jajodia2026beforeyou} presents a set of data and an analysis framework for evaluating cyber security incident analysis.  CORTEX \cite{wei2025cortex} applies similar tradecraft to provide step-by-step analyst actions paired with tool outputs to enable the evaluation of multi-agent alert triage workflows. Finally, CyberGym \cite{cybergym2026} presents an open, large-scale benchmark featuring $1507$ real-world vulnerabilities across $188$ software projects, aiming to address the limitations of static, small-scale cybersecurity evaluations. Although these efforts are a massive step forward for the AI security arena, it is important to note that the threat landscape is adversarial and is constantly evolving. As such, a benchmark calibrated to a static point-in-time set of attacks will almost certainly not reflect the attack surface in the future, thus becoming operationally stale.

All of the hard work on benchmarks still leaves many scenarios with limited labeled data, or at least limited up to date labeled data, and a need to test a learning harness to see if it learns properly. For example, a large number of AI SOC scenarios have limited labels and rapidly evolving labels and data, but testing such systems and their ability to learn remains critical to their usefulness. In addition, \cite{happe2025benchmarking} survey test beds and metrics for testing LLMs for offensive security, and show that test beds and the system as a whole are key to the results of an evaluation, which suggests that rather than a set of fixed test examples, we need a way to test learning harnesses end to end in their native context with native data. 

\subsection{Agent Harness Evaluation}
Evaluating an agentic harness presents even greater challenges than assessing the underlying model. Unlike standard language model benchmarks, agents function within dynamic and interactive environments, where a single held-out test set cannot capture the full spectrum of relevant behaviors, \cite{mohammadi2025evaluation}. \cite{yehudai2025survey} divide the evaluation challenge into five perspectives: core agentic capabilities such as planning and tool use, application-specific benchmarks, generalist agent evaluation, benchmark design dimensions, and developer-facing frameworks. They identify persistent gaps across all five areas, particularly in cost-efficiency, safety, and cross-task robustness \cite{yehudai2025survey}. As tasks increase in length and complexity, these gaps become more pronounced. Reliability increasingly depends on supporting infrastructure, including memory stores, orchestration loops, and retrieval pipelines, rather than solely on the model weights \cite{ameng2026harness}. This situation leads to a conflation problem: task accuracy on a labeled test set reflects the combined performance of both the model and the harness, making it difficult to isolate the harness's specific contribution, and difficult to predict how accuracy might evolve as labels and their availability evolve over time.

To reduce the dependency on expensive annotated data, researchers in the LETToT \cite{qi2025lettot} paper leveraged expert reasoning chains to map out logical steps for planning or evaluation. Likewise, GEPA \cite{agrawal2026gepa} uses LLM judges as proxies for ground-truth. However, both of these efforts rely on the implicit assumption that the reference signal is consistently trustworthy enough to serve as a stand-in for ``gold" labels. Our framework makes that assumption explicit and attempts to ground it empirically by designating a frontier model as a teacher. If the student harness begins to mirror the teacher's predictions, we can say the harness has added value without the hard requirement of ground-truth labels. This differs from classical distillation approaches like Hinton's \cite{hinton2015distilling}, where distillation fine-tunes the student based on teacher output, baked directly into the model weights, and therefore not inspectable. Our approach aims to have a reflection model author a natural-language lesson for each disagreement between a teacher and a student. At inference time, these lessons can be retrieved by finding a previous case where the lesson is about a similar topic to the current need for inference and injecting them into the student's prompt; the model's weights are never modified.

\subsection{Agent Memory Evaluation}
Much agent learning and memory is powered by retrieval-augmented generation (RAG), which enables rapid adaptation by improving model behavior through the retrieval of external context during inference. This methodology is particularly beneficial in cybersecurity, where adversarial environments prevail. Defensive platforms must address a constantly evolving array of attack patterns and local analyst feedback, which complicates efforts to keep training cycles up to date.

Agentic memory systems build upon this concept by storing past errors, feedback, and corrections as reusable lessons or memories. Prior research, such as Reflexion \cite{shinn2023reflexion}, MemPrompt \cite{madaan2022memprompt}, Learning to Repair \cite{tandon2022learning}, and MemoryBank \cite{zhong2024memorybank}, demonstrates this broader approach: systems iteratively refine future behavior through structured lesson retrieval instead of model weight updates. Within this framework, an effective lesson should both clarify the reasons for misclassifications and improve the agent’s subsequent performance. This functionality enables the capture of analyst corrections and their retrieval for structurally and semantically similar messages.

In security, classification tasks diverge from traditional RAG applications. Conventional systems typically assume that most retrieved context will support decision-making and, therefore, optimize for ranking the top-k passages. In contrast, security classification scenarios often require abstention, as there may not be relevant lessons in every instance. Returning a top-k result in such cases may reduce efficacy and introduce regressions. As a result, a memory system should implement an absolute relevance threshold rather than relying exclusively on relative ranking. Evaluation should focus on whether corrections are retained and effectively applied to future cases. A lesson derived from one message should improve performance on a second, semantically similar message that was not available during the initial lesson creation. 

The evaluation literature for RAG and memory-based systems emphasizes a limited set of metrics. Traditional information retrieval evaluations assess retrievers using metrics such as \textit{Precision@k}, \textit{Recall@k}, \textit{Mean Reciprocal Rank}, and \textit{NDCG@k}, which quantify the frequency with which relevant documents appear near the top of ranked lists. These metrics are widely used in BEIR \cite{thakur2021beir} and TREC \cite{voorhees2005trec} benchmarks. RAGAS \cite{es2024ragas} extends this by evaluating RAG outputs with context recall and context relevance, which assess whether the retrieved context provides the necessary material to support the final output. Self-RAG \cite{asai2024selfrag} separates retrieval and generation, comparing live retrieval against an oracle to determine whether failures arise from a missing or improperly used context. ARES \cite{saadfalcon2024ares} addresses scalability by employing a large language model as a judge, validated against the output of human annotators, to evaluate context quality. Memory-based approaches such as MemPrompt, Reflexion, and MemoryBank introduce persistence-based metrics to determine whether feedback enhances performance on future, related samples, whether stored lessons generalize beyond the original task, and whether retrieval becomes biased toward newer lessons due to recency rather than semantically similar memories.

Collectively, these lines of research indicate that the application of lesson-augmented methods in cybersecurity requires an evaluation strategy that spans three vectors: retrieval quality, persistence of quality feedback, and abstention under uncertainty. Thus, the main retrieval challenge is both ranking candidate lessons and deciding if a lesson is similar enough to act upon. In the case of security, these are often scenarios where labels are more expensive and unclear than the labels for correct results for the full end to end system. 

\section{Experiment Design}

The setting we ultimately care about is a production operational security model. Here, most people are already running the strongest model available, and care to steer its behavior over time with only sparse, high-precision human feedback, the kind that arrives only in a high-severity escalation. We can't evaluate this regime directly as there's no gold benchmark. And the absence is the entire problem. We therefore aim to design our experiments as a measurable proxy for the regime and hold as much of the scenario constant as we can. We can view the human correcting a model on escalated cases as a high precision indicator of truth with a large gap over the model it is aiming to correct. So as a proxy, we try and substitute a stronger model teacher for the human and preserve the rest. We aim for corrections to still be sparse, high precision, and delivered over time in rounds.  We measure  each Harness across many student-teacher pairs, so we can evaluate a Harness's contribution semi-independently of the model piloting it.  We conjecture that if the proxy metrics work in this measurable regime, they should (since so much of the correction methodology is mirrored) hold for the human-frontier regime we cannot directly benchmark.

We perform two sets of experiments. The first set is to show that student uplift when sparsely learning from a teacher as measured on teacher pseudo-labels is correlated with golden dataset uplift under the same learning conditions. We illustrate this with several security related tasks. The second set of experiment confirms that  using an LLM as a judge strategy is not a good measurement in this case, if the LLM being used in the learning harness is of a similar caliber to the judge, and thus this means of measurement is not likely to yield valuable data. 

\subsection{Continual Learning Methodology}

Our central methodological claim is that the accuracy uplift a learning harness
delivers can be measured against a stronger model's pseudo-labels in place of the
true labels. To test this, we set up a continual learning procedure that never
uses gold labels except at the very end, for evaluation.

We use three models per task: a \emph{teacher} (the stronger model, run without a
harness), a \emph{student} (the weaker model, run without a harness), and
\emph{student*} (the same student equipped with the learning harness). We first
have the teacher predict every item in the benchmark and treat these predictions
as our pseudo-labels. We then partition the dataset into five splits: four for
successive rounds of continual learning, and a held-out fifth split that serves as
the test set.

In each of the four rounds, student* predicts the items in that round's split. To
mimic how a human analyst corrects a model in security settings, we then draw a
random sample of up to five ``high-impact'' misclassifications from that round and
feed them back to student* as corrections. Student*'s behavior changes from round
to round based on the corrections it has accumulated. After all four rounds,
student* has seen no more than 20 corrections out of hundreds of data points,
deliberately reflecting the sparse, sporadic supervision available in practice.

Finally, we evaluate on the never-seen test set (split 5) using both the raw
student and student*. Let $F$ denote the benchmark's chosen metric (it varies by
task). We score student* and student against two different reference sets, the
teacher's pseudo-labels and the gold labels, and measure the harness uplift under
each:
\begin{align}
\Delta G &= F(y_\text{gold},\, \hat{y}_\text{student*})
          - F(y_\text{gold},\, \hat{y}_\text{student}), \\
\Delta T &= F(y_\text{teacher},\, \hat{y}_\text{student*})
          - F(y_\text{teacher},\, \hat{y}_\text{student}).
\end{align}
$\Delta G$ and $\Delta T$ are computed on identical models and identical test
items; they differ only in the reference labels (gold vs.\ pseudo). Their
relationship therefore measures whether teacher pseudo-labels rank harness uplift
the same way gold labels do. Notice that gold labels are never used at any point
other than this final evaluation.

We repeat this procedure across multiple harness types and numbers of corrections,
and measure the correlation between $\Delta G$ and $\Delta T$ for the task at hand,
reporting both the Pearson coefficient $r$ and the Spearman rank coefficient
$\rho$. We then repeat the whole process across several security tasks for which
published benchmarks provide gold labels, to confirm that the correlation holds
broadly rather than for any single task.

\subsection{Model and Security Task Selection}
We evaluate a full factorial of three security tasks, three model families, four harness configurations, and two supervision conditions.

The three tasks span an array of label granularity and difficulty:
\begin{itemize}

\item \textbf{PhishFuzzer}: a three-way email classification (phishing, spam, benign).

\item \textbf{CTIBench}: root-cause attribution, mapping each CVE description to its underlying CWE.

\item \textbf{ATT\&CK tactic classification}: a fine-grained task derived from CTIBench by tagging behaviors at the technique level and joining against the MITRE ATT\&CK matrix to recover tactic labels.

\end{itemize}

We chose these tasks as we wanted to limit our methodology to the security operational setting and these were 3 rare security datasets with gold labels spanning a difficulty/granularity gradient. This lets us compute $\Delta G$ and test whether the proxy holds as the task gets harder and across different security domains. ASA itself has no gold, which is the whole problem, so we chose these datasets to validate where gold exists and treat ASA as the deployment the proxy is ultimately for.

We choose 3 model families (one small and one large from each family) to be our students and teacher models. We chose these as they are 3 architecturally distinct families at two scales each, so you can form same-family pairs with a verified gap (Fig 1), as well as test out what would happen if we had cross-family students and teachers.

Across each task we run three model families (Gemini, GPT, and Qwen) under four harnesses: a no-learning \textbf{baseline} and three learning harnesses, \textbf{Memory}, \textbf{Lesson}, and \textbf{Few-shot}.
Each configuration is tested under two supervision conditions. \emph{Proper} supervision draws labels from the teacher model or the golden dataset; \emph{sabotaged} supervision deliberately injects incorrect labels. Sabotage is our negative control: a faithful harness must lose accuracy under corrupted supervision, ruling out the possibility that observed gains are evaluation artifacts rather than genuine learning.

For each task, we used a baseline harness with no learning capability and three different learning harnesses on each of three model families: Gemini 2.5 Flash \& Pro, GPT 4.1 Nano \& 4.1, and Qwen-7B \& 72B  \begin{itemize}
    \item \textbf{Memory:} Each retrieved correction is rendered as a record of a specific prior error. This includes a short excerpt of the similar past input, the student's \emph{incorrect prediction}, and the correction label. This is basically the error-memory approach of MemPrompt \cite{madaan2022memprompt}, Reflexion \cite{shinn2023reflexion}, and MemoryBank \cite{zhong2024memorybank}.

    \item \textbf{Fewshot:} Each retrieved correction is rendered as a labeled exemplar (input excerpt $\rightarrow$ correction label), presented as a standard few-shot example for the student to generalize from. Unlike Memory, the past item is shown as a canonical solved instance and the previous error is not presented.

    \item \textbf{Lesson:} Rather than storing the raw example, a separate instance of the student model reflects on the error and summarizes each mistake into a short discriminative \emph{lesson}. This is a slightly more general rule of thumb for telling the confused label apart from the correct one, with explicit ``apply only when" and ``do not apply when" cues. Retrieved lessons are injected as hypotheses and the student decides from the input's own content whether to apply a lesson. 
\end{itemize}

The \textbf{Lesson} harness is a simplified version of the harness that we have hypothesized to work for improving and personalizing ASA from customer specific feedback. Its value concentrates on the hard personalized cases, the borderline emails where the model is confused and small differences in wording or intent flip the correct label.

One concern is that $\Delta T$ might track $\Delta G$ only because teacher and student share a family, and thus have correlated errors. To rule out that this proxy only works for in-family pairs, we also test cross-family pairs, selecting from Figure 1 only those where the teacher beats the student on the base task. Without the assumptions of the same training data and architecture techniques, their errors are more decorrelated, and we know that its the capability gap driving the growth, not the same-family correlations. This is also more closely resembles our eventual target regime: a human teacher over a frontier student is the limiting large-gap, not a shared model family case. 

\section{Experiment Results}

\subsection{Teacher Alignment Correlated with Ground Truth Alignment}

\begin{figure}
    \centering
    \includegraphics[width=1\linewidth]{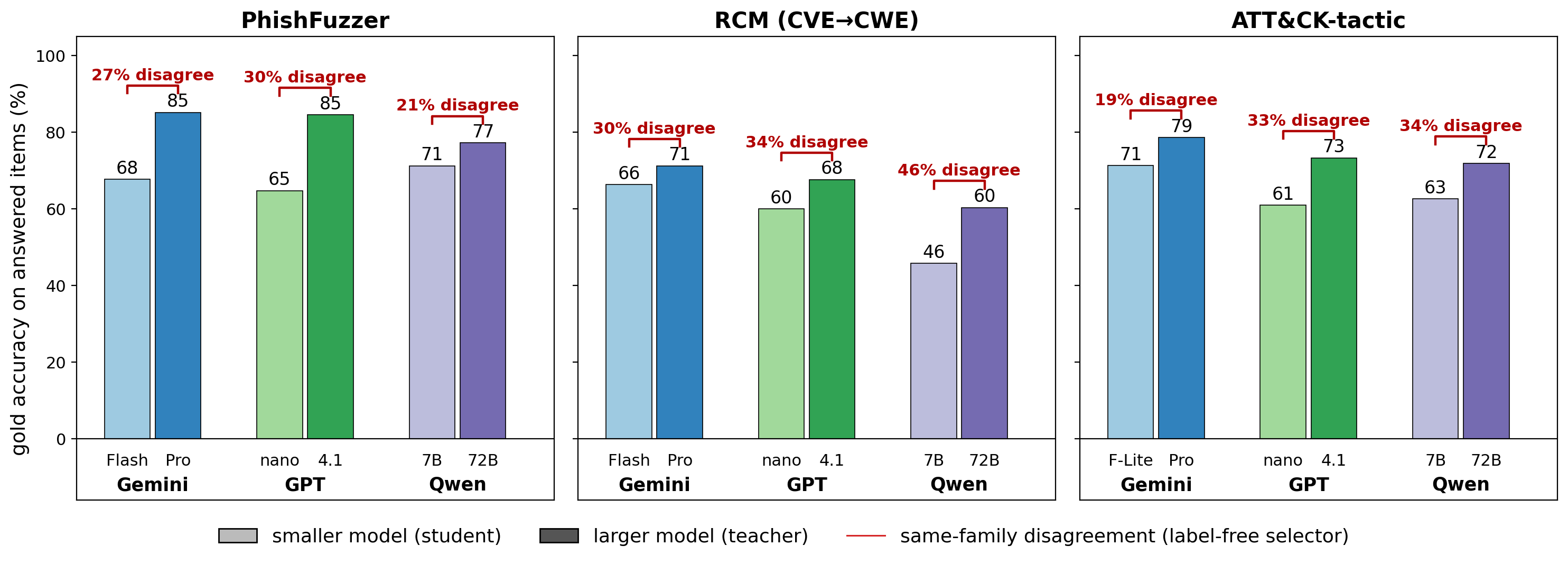}
    \caption{In baseline Performance of LLMs tested on the benchmarks, within every family, the larger model out performs the smaller model on each of the three tasks. This is not surprising, as it is the same as saying that the scaling hypothesis holds. }
    \label{fig:Baseline}
\end{figure}

We begin by checking that the scaling hypothesis holds as expected for the tasks and model families. In figure \ref{fig:Baseline}, we see that, as expected, for each model family the baseline performance of a larger teacher model is higher than the baseline performance of a smaller student without learning. In the case of these tasks, it is also true that the larger teacher model from each family is stronger than the student model from each other family, though the scaling hypothesis alone does not imply this is true.

\begin{figure}
    \centering
    \includegraphics[width=0.75\linewidth]{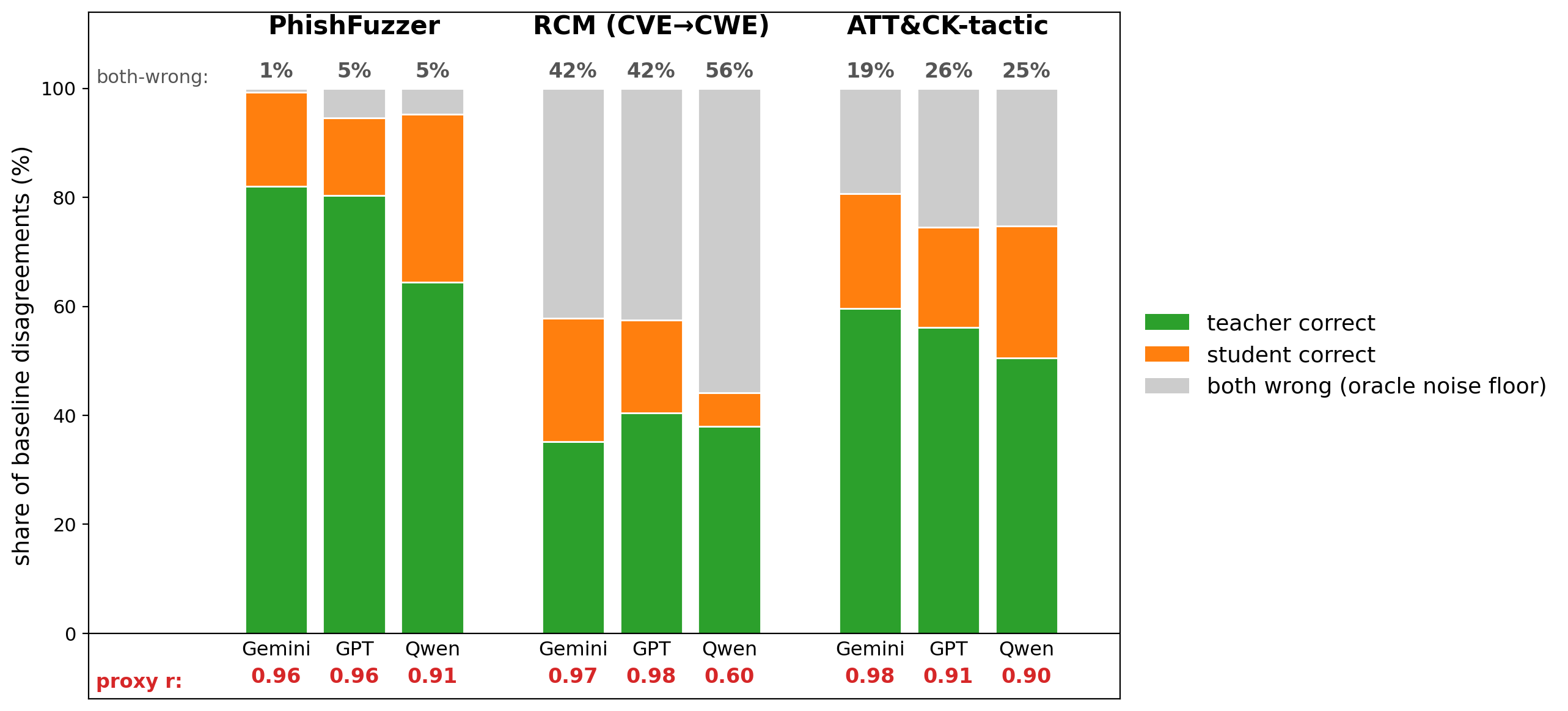}
    \caption{In cases where the student and teacher disagree, the teacher (scaled up version of the student in the same family) is correct far more than the student on hidden gold label. This supports what is expected from the scaling hypothesis}
    \label{fig:StudentTeacherDisagree}
\end{figure}

In addition, we check that when the student and teacher are not in agreement, the teacher is more often correct than the student, something we expect would be true even in the absence of labels, shown in \ref{fig:StudentTeacherDisagree}. 

Critically, we show that in the case of all pairs of student and teacher models, there is a strong correlation between additional learning from a teacher and additional performance on the golden labeled dataset. In addition, for most task and model pairs, the sign of the uplift from golden labels and teacher labels is the same, indicating that we could tell if a harness will improve a model's performance on a task with proxy labels from stronger larger models in the same family. 
\begin{figure}
    \centering
    \includegraphics[width=0.75\linewidth]{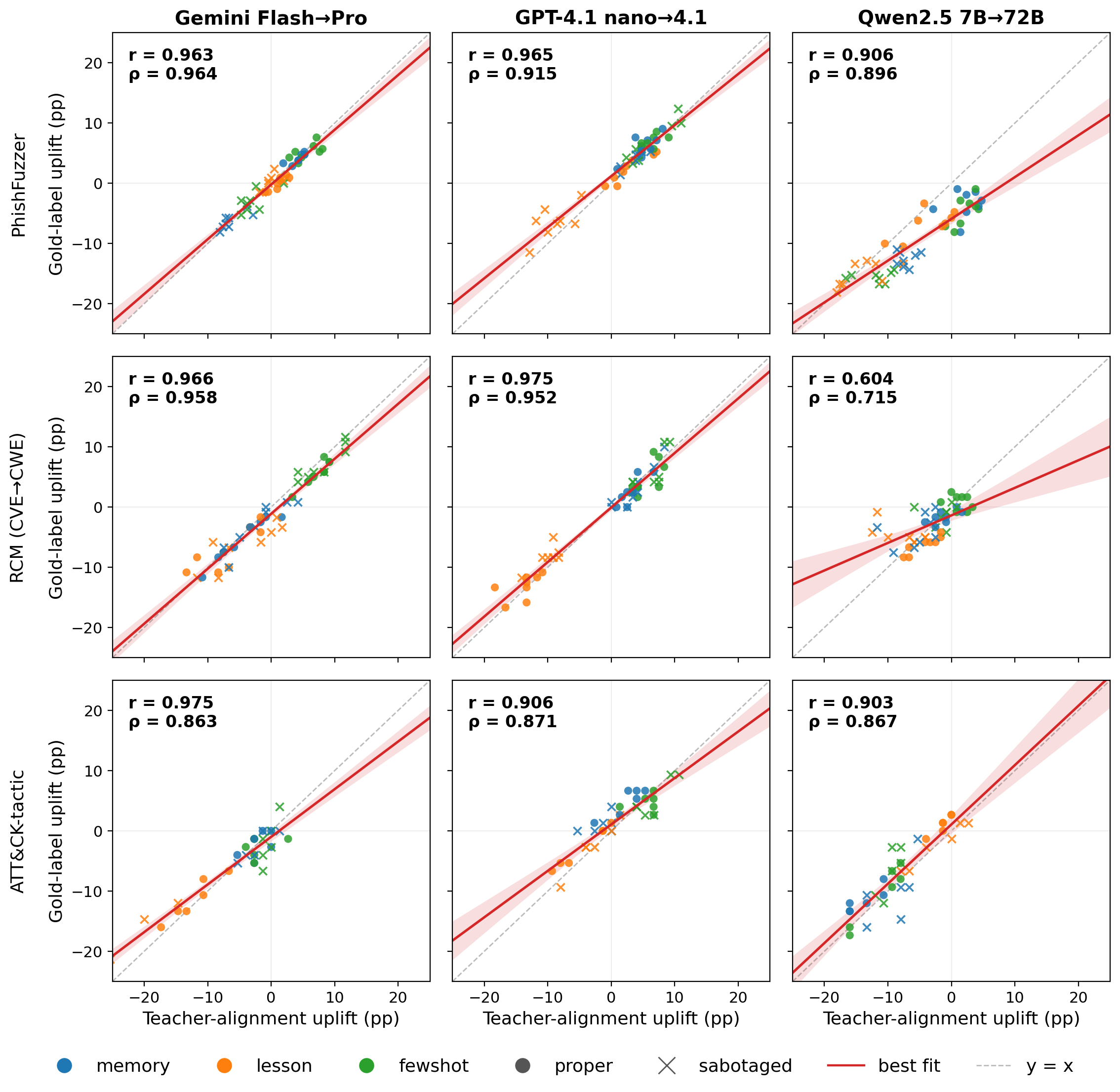}
    \caption{Held-out teacher alignment shows strong positive correlation with gold label uplift across model families and tasks}
    \label{fig:SameFamilyUplift}
\end{figure}

Further, this correlation does not depend on the teacher and student sharing a family. When we pair students and teachers from different families, so long as the teacher outperforms the student on the base task, the $\Delta T -\Delta G$ correlation persists (Figure 4), and is comparable to the strength of the same family pairs for the same task. Because cross-family pairs are more de-correlated on architecture and training data, it seems to indicate what governs the proxy's validity is a capabilities gap between student and teacher, not specifically the model family and model size within the family. The proxy seems to hold wherever a genuine, positive, capability gap exists. The cases where the correlation weakens (visibly Qwen on RCM in Figure 3 and the qwen-7B→Gemini-Pro pair on ATT\&CK-tactic in Figure 4) are precisely those where the gap is smallest. And we often see the strongest correlation where the capability gap between student and teacher is the largest.

\begin{figure}
    \centering
    \includegraphics[width=0.75\linewidth]{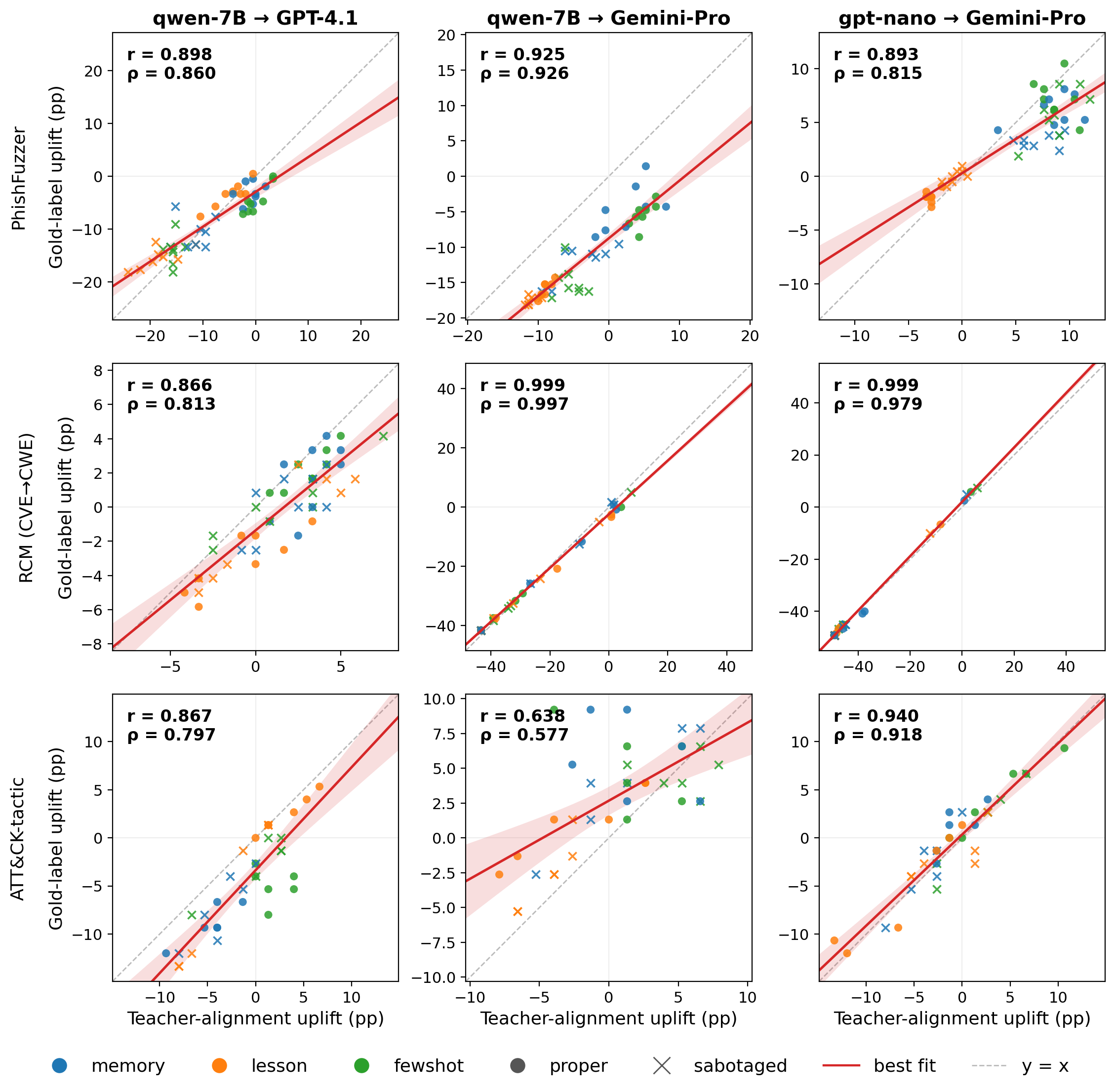}
    \caption{Held-out teacher alignment shows strong positive correlation with gold label uplift even in cross-family student-teacher pairs where teacher ranked higher than student}
    \label{fig:CrossFamilyUplift}
\end{figure}

\begin{figure}
    \centering
    \includegraphics[width=1\linewidth]{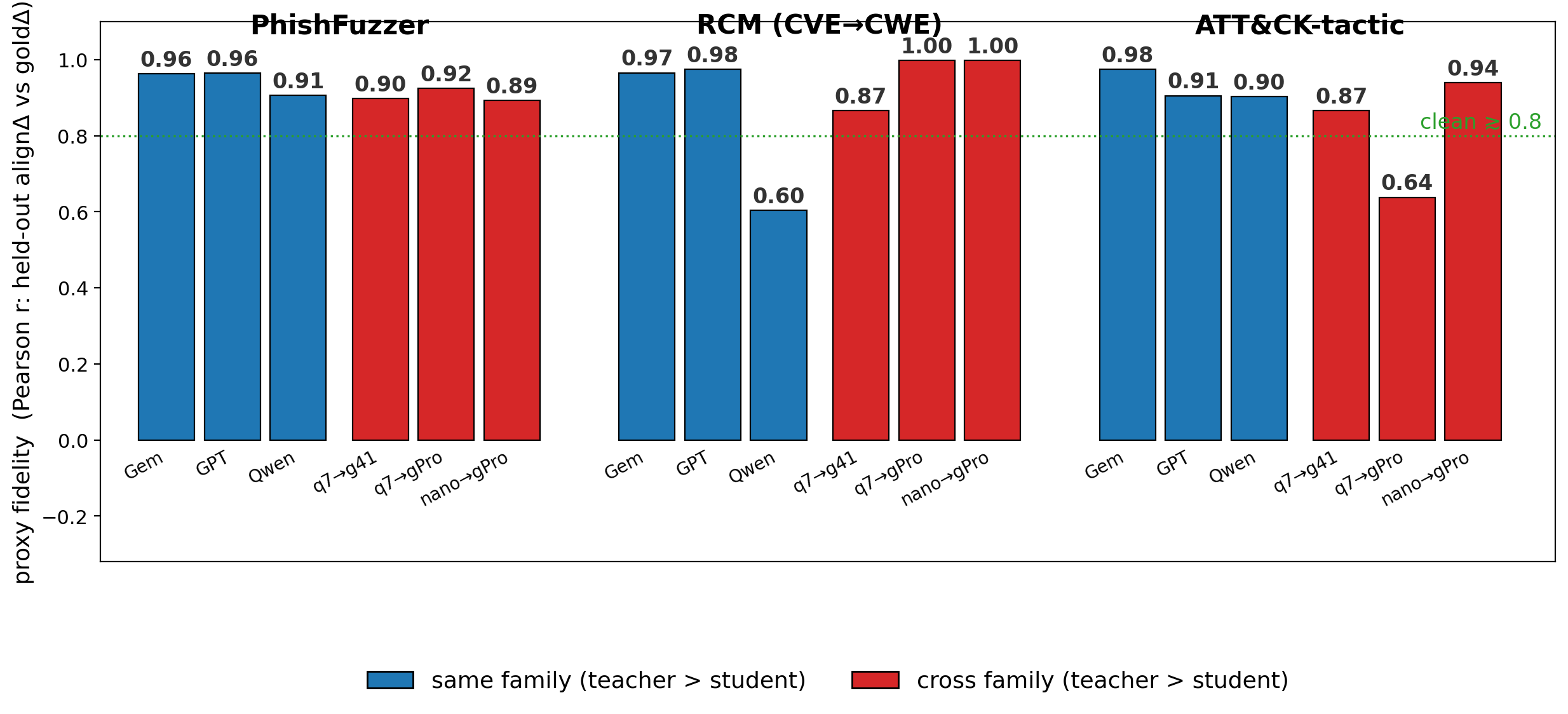}
    \caption{Cross family correlation is generally comparable to within family correlation so long as teacher $>$ student on base task performance.}
    \label{fig:CrossVsWithin}
\end{figure}

We compare cross family correlation for a given task with within family correlation in figure \ref{fig:CrossVsWithin}, showing comparable results. 

\subsection{Inefficacy of LLM-as-Judge}

In contrast to teacher alignment, using an LLM as judge to determine whether outputs of a model equipped with a learning harness, are better than the base model is largely unreliable if the judge is of a similar stength to the base model. Models prefer their own opinions over those of an even better model when presented with a choice. This means that if the desire is to use a frontier model in production for a learning harness, one cannot use a frontier model as a means of measuring if a learning harness works. 

We again evaluated three different model families Gemini, GPT, and Qwen, with three different harnesses memory, lesson, and few-shot on the \textbf{PhishFuzzer} email classification dataset described above. In addition to teacher alignment metrics, we present judge preference, and judge alignment with ground truth. 

We create a judge as a copy of the student for each of three model families. For each email it sees the two candidate labels (with-harness vs baseline) and is instructed to pick the more accurate label; without direct access to the ground truth. Each pairwise comparison is judged twice, with answers switching A/B positions, in order to disentangle model preference from positional bias. We pool the performance across 3 different harnesses since each individually exhibits a similar pattern.

\begin{table}[ht]
  \centering
  \begin{tabular}{lccc}
    \toprule
    \textbf{Family}                  & \textbf{Positional} & \textbf{Decisive verdicts}      & \textbf{Reliability vs.\ gold} \\
    \textbf{(same-strength judge)} & \textbf{flip rate}  & \textbf{(\% pref.\ no-harness)} & \textbf{($\geq$1 model correct)} \\
    \midrule
    GPT-4.1-nano & $96\%$ & $0\%$  & $0\%$  \\
    Gemini-flash & $74\%$ & $40\%$ & $7\%$  \\
    Qwen-7b      & $0\%$  & $74\%$ & $56\%$ \\
    \midrule
    Pooled       & $50\%$ & $69\%$ & $48\%$ \\
    \bottomrule
  \end{tabular}
  \caption{Student-strength judge comparisons - \textbf{Baseline vs with Harness}. Decisive-verdict and reliability
           columns are conditioned on decisive verdicts and on at least one
           model being correct, respectively.}
\end{table}

We notably found the following behavior from the same-strength judge:

\begin{enumerate}
    \item \textbf{Positional Bias} Two of three judges flip their verdict when the same two answers swap positions - exhibiting the well documented position bias of favoring the first presented option, see \cite{zheng2023judgingllmasajudgemtbenchchatbot}. We suspect that when it is unclear which option is better the judge resorts to positional biasing.
    \item \textbf{Anti-harness preference} Where the judge does commit to a stable verdict, it prefers the un-harnessed baseline a majority of the time - directly contradicting the helpfulness of the harness we see against teacher alignment and ground truth.
    \item \textbf{Poor reliability against ground truth} Most importantly, on cases where at least one (of baseline and learning harness model) got the correct classification, the judge's preference did not have high alignment with the ground truth classification for the emails, in other words, if either the harness or the baseline is wrong, the judge isn't likely to be a good judge of truth. 
\end{enumerate}

As another illustrative example, on the same task and same model families, we ask a student level judge to compare answers between its own vs its teacher models. We do this to determine if student's are able to recognize answers better than their own when there is a large difference in performance.

\begin{table}[ht]
  \centering
  \begin{tabular}{lcccc}
    \toprule
    \textbf{Pair}            & \textbf{Positional} & \textbf{Prefers teacher} & \textbf{Prefers its} & \textbf{Reliability} \\
    \textbf{(student judge)} & \textbf{flip rate}  & \textbf{(of decisive)}   & \textbf{own answer}  & \textbf{vs.\ gold} \\
    \midrule
    GPT-4.1-nano & $7\%$         & $5\%$  & $95\%$ & $17\%$ \\
    Gemini-flash & $26\%$        & $26\%$ & $74\%$ & $38\%$ \\
    Qwen-7b      & $18\%$        & $48\%$ & $52\%$ & $26\%$ \\
    \midrule
    Pooled       & $15\%$  & $22\%$ & $78\%$ & $25\%$ \\
    \bottomrule
  \end{tabular}
  \caption{Student-judge comparisons. The ``prefers teacher'' column is
           conditioned on decisive verdicts; the reliability column is
           conditioned on at least one model being correct.}
\end{table}

Pooled across families, the same-strength judge reaches 48\% reliability against gold — no better than a coin flip on a binary choice — and the student-judge falls to 25\%, below chance, meaning it systematically prefers its own answer to the teacher's better one. Here the judge reveals a strong self-preference bias. A student judge has no ability to recognize a better model answering above its own capability. In the unlabeled case where at the frontier our only option is a student judge, we cannot rely on that judge to see improvements beyond its own base capabilities.

\section{Conclusions}

\subsection{Transfer to Operational Security Regime}
So do these results transfer to the operational security regime that actually
exists today in practice? We argue that they do. Our results already show that
the proxy's validity is not tied to model size and is not limited to pairs from
the same family. It is governed by two things: the student--teacher capability
gap on the task, and the precision of the corrections the teacher supplies.
Critically, it does not matter whether those high-precision corrections come from
a model or from a human, because the harness only ever interacts with the
corrections themselves, never with the teacher in any other way. The teacher is
just a source of high-precision labels, and a human analyst is another.

The Scaling Hypothesis is what lets a practitioner find a valid student-teacher pair in practice. A substantially larger model from the same family is very likely the stronger one on most tasks, and so on the task at hand too. A human reviewing an escalated security case is the extreme version of that same student-teacher relationship: a higher-precision predictor of the truth with a large gap over the production model. Because our experiment design matched the operational correction protocol with sparse, high-precision, corrections delivered over a few rounds, the validated setting already resembles what our harness is sensitive to. 

For the operational scenario with no gold labels, if a specific (harness, task) pair, say lesson sketching for the ASA task, shows high proxy uplift ($\Delta T$) across multiple model families, it stands to reason it adds value independent of the piloting model, and would transfer to the human-frontier regime. This conjecture seems to follow logically from our experiment results, and we predict that it is a valid way to measure the harness capability in operational security settings. It would be falsified if we saw consistent cases where teacher-relative and gold-relative uplift decoupled despite a large capability gap, if a direct test of the human case against held-out gold showed teacher-relative uplift de-correlating.

\subsection{Limitations}
 The proxy's two dependencies are its failure modes. We see that when the capability gap is small, or when both student and teacher are extremely bad at the task, the correlation, despite still being present, drops significantly. We see this in our results for Qwen on RCM (figure \ref{fig:StudentTeacherDisagree}). In practice, it is hard to measure these prerequisite dependencies without labels. However, we can still mitigate these by using proxies for these dependencies. Against the student-teacher capability gap, we can use the scaling hypothesis to our advantage. A substantially larger model from the same family is likely stronger than a smaller one on most tasks. We can run a large sample of unlabeled tasks through this larger model for pseudo-labels, and can do the same for a much smaller model in the same family. If the disagreement is large, it suggests that the gap is large, because for each disagreement between small-student and large-teacher, the teacher is more likely to be correct. The idea is to test the harnesses against multiple model families where these disagreement rates are large. Against the situation where both student and teacher are extremely bad at the task, such terrible performance from the teacher model can be spotted with just a handful of labeled samples, and still doesn't require a full sample. The principle we have noticed here is that the teacher needs to be correct a clear majority of the time, a threshold below which no security team would knowingly deploy an agent in the first place, which can often by checked with a small number of labeled samples. 

\subsection{Future work}
The obvious extension to this work is to validate the human-frontier conjecture against a real, gold labeled, operational security dataset. This would require a lot of effort to get gold labels on a non-toy task, and lots of human labeling which would take a lot of effort, but even a small dataset may be enough to provide evidence that the conjecture holds. In addition, our current framework seems to work a bit like operational distillation, which opens up future exploration. This is a good avenue for exploring cost efficiencies without long term effectiveness sacrifices. It also opens up the question of baking in accumulated corrections back into the system prompts or even directly into the weights via adapters like LoRA, and comparing how these different versions of harnesses perform on the proxy metrics.

\subsection{Final Conclusion}
Most importantly, this work gives security operators a way to tell whether a harness helps at all, as well as compare two harnesses, before committing one to production in a far more rigorous way, with a strong basis for it working on human corrections to frontier models. 
\clearpage
\bibliography{references}

\end{document}